# LudVision
# Remote Detection of Exotic Invasive Aquatic Floral Species using Drone-Mounted Multispectral Data

**António J. Abreu**
NOVA LINCS, Universidade da Beira Interior
R. Marquês de Ávila e Bolama Covilhã, 6201-001, Portugal
Zirak srl – Information Technology
Via S.Agostino 6/A Mondovì (CN), 12084, Italy
antonio.abreu@ubi.pt

**Luís A. Alexandre**
NOVA LINCS, Universidade da Beira Interior
R. Marquês de Ávila e Bolama Covilhã, 6201-001, Portugal

**João A. Santos**
Zirak srl – Information Technology
Via S.Agostino 6/A Mondovì (CN), 12084, Italy

**Filippo Basso**
ECOATIVO - Associação de Proteção e Conservação da Natureza
Rua dos Três Lagares Fundão, 6230-421, Portugal

## Abstract

Remote sensing is the process of detecting and monitoring the physical characteristics of an area by measuring its reflected and emitted radiation at a distance. It is being broadly used to monitor ecosystems, mainly for their preservation. Ever-growing reports of invasive species have affected the natural balance of ecosystems. Exotic invasive species have a critical impact when introduced into new ecosystems and may lead to the extinction of native species. In this study, we focus on Ludwigia peploides, considered by the European Union as an aquatic invasive species. Its presence can negatively impact the surrounding ecosystem and human activities such as agriculture, fishing, and navigation. Our goal was to develop a method to identify the presence of the species. We used images collected by a drone-mounted multispectral sensor to achieve this, creating our LudVision data set. To identify the targeted species on the collected images, we propose a new method for detecting Ludwigia p. in multispectral images. The method is based on existing state-of-the-art semantic segmentation methods modified to handle multispectral data. The proposed method achieved a producer's accuracy of 79.9% and a user's accuracy of 95.5%.

## 1 Introduction

Exotic invasive species have a critical impact when introduced into new ecosystems, and there is a growing global concern due to their negative ecological and economic ramifications. The introduction of invasive plants often results in the extinction of native plants and a reduction in biodiversity, either by outcompeting or hybridizing with native species [1]. This is especially the case in aquatic environments and for aquatic plants [2]. Under most scenarios, invasive plants arrive without their co-evolved competitors or parasites, allowing them to spread rapidly, replacing native plants without assuming their ecological roles.

Invasives in wetlands have many adverse effects within and across trophic levels and significantly reduce biodiversity [3, 4]. Many invasives may directly compete with other species by secreting allelopathic chemicals that reduce germination and seedling survival or by changing light accessibility [5, 3, 6]. Invasives may also significantly impact invertebrate distribution, diversity, and abundance; induce anoxic conditions detrimental to fish and other aquatic life [5, 7]; and act as barriers to fish movement [6, 4]. They also reduce open water habitats for water birds and other wildlife [4].

RS image analysis is increasingly being used to map invasive plant species. The resulting distribution maps can be used to target the management of early infestations and to model future invasion risks [8, 1]. Remote identification of invasive plants based on differences in spectral signatures is the most common approach, typically using hyperspectral data [9]. Advances in Unmanned Aerial Vehicle (UAV) and sensor miniaturization are enabling higher spatial resolution species mapping, which is promising for early detection of invasions before they spread over larger areas [10].

Our goal is to monitor the spread of Ludwigia peploides, specifically in the Reservoir of the Toulica Dam (Zebreira, Portugal), located in the hydrographic basin of the Aravil river, a tributary of the Tagus, where it was detected in 2020. Ludwigia peploides is a species natural to South America that invades rivers, ponds, and rice fields. It can grow in deep waters, as a fully or partially submerged plant, and form floating mantles. When this happens, it prevents the entry of light affecting submerged species and blocking the water lines, affecting navigation, fishing, and recreational use. It competes for space by eliminating native species and producing substances that inhibit the germination and growth of other species. It reproduces vegetatively through stem fragmentation but also seeds. It is an invasive species that raises concern at the European level (EU Reg. 1143 [11]) and is included in the Portuguese legislation list of invasive species (DL 92/2019 of 10/07 [12]). Although our focus was on this particular location, the developed detection method is general and can be used with data from other locations.

As such, we aimed to develop a system for the remote detection of the Ludwigia peploides. To do so, we first created the LudVision data set. It contains various images of Ludwigia peploides taken at several altitudes, times of day, and atmospheric conditions. This was to ensure that our data set was as representative as possible. We also propose a new method for detecting Ludwigia p. in multispectral images. The method is based on existing state-of-the-art semantic segmentation methods that were modified to handle multispectral data.

Since remote sensing applications typically rely on specialized and expensive equipment, we set on using a general application sensor and platform instead of the highly specialized and expensive equipment traditionally used in RS applications. We wanted to prove that using semantic segmentation models could create high-reliability maps with available application equipment. Our LudVision data set will be provided for public use.

## 2 Related Work

Over the years, remote sensing has received much attention, especially for the detection of invasive species [13, 1, 14, 15, 16, 17, 18, 19]. Specifically, there has been an increasing interest in detecting and monitoring aquatic invasive species [10, 16, 17, 7] due to the severe negative impacts they can have on ecosystems. One prime example of invasive aquatic species are water hyacinths and ludwigia [10, 7, 16, 17]. Due to improvements in sensor technology and the availability of UAVs, the use of unmanned aircraft has gained a lot of attention [10, 20, 18, 19]. The low costs of acquisition and operation, paired with the ability to be deployed almost instantly without significant planning, have made them a viable alternative compared to more expensive data like Manned Aerial Vehicles or satellites. They are beneficial when high resolution and frequent acquisition is a requirement. The only major downside of drone-mounted sensors is the limited capability to cover broad areas. One of the main ways species can be detected and differentiated is by performing spectroscopy analysis [14, 13, 1, 17, 15]. Image spectroscopy relies on the fact that each species has a unique spectral signature which can be used to identify a species. Multispectral or preferably hyperspectral imagery is needed to extract the spectral signature of a species.

There has been vast interest and research done regarding semantic segmentation, with multiple interpretations and adaptations. Given that our problem needs to be able to keep high-resolution representations, we analyzed a few subsets of models specifically designed for high-resolution representations [21, 22, 23, 24, 25, 26, 27]. The concepts applied in the studied models were used to develop our proposed model.

Despite the extensive research and work performed in both remote sensing and semantic segmentation, we could not identify any work done that applied semantic segmentation models to remote sensing. Thus we propose a new semantic segmentation model as the classifier to detect Ludwigia peploides.

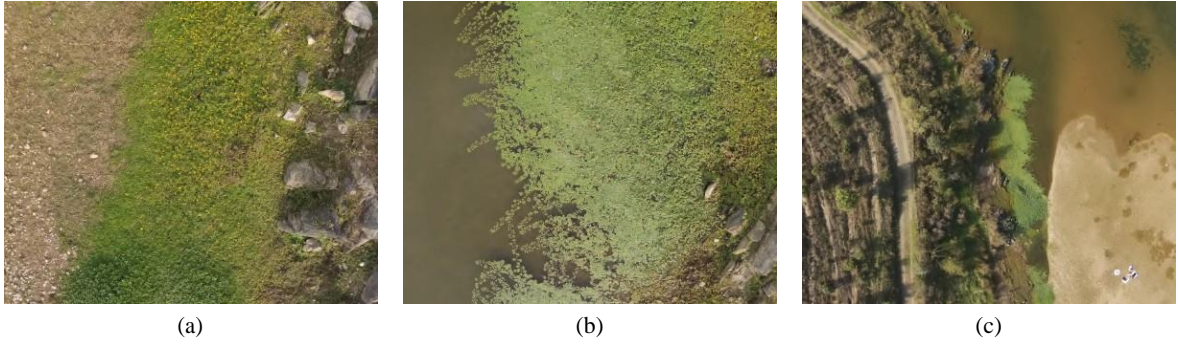

(a)                       (b)                      (c)

Figure 1: Images from the Ludwigia peploides collected at the Reservoir of the Toulica Dam at different altitudes. (a) 15 m, (b) 40 m, and (c) 70 m.

## 3 Methods

### 3.1 Study Site and Targeted Species

The study site is the Reservoir of the Toulica Dam (Zebreira, Portugal), located in the hydrographic basin of the Aravil river, a tributary of the Tagus. In 2020 the IAS Ludwigia peploides was spotted in the northeast part of the reservoir and had since spread, forming three big mantles. Recently it started spreading and forming small patches southwest of the initial infestation site. Ludwigia peploides is a species natural to South America that invades rivers, ponds, and rice fields. It can grow in deep waters, as a fully or partially submerged plant, and form floating mantles. When this happens, it prevents the entry of light affecting submerged species and blocking the water lines, affecting navigation, fishing, and recreational use. It competes for space by eliminating native species and producing substances that inhibit the germination and growth of other species. It reproduces vegetatively through stem fragmentation but also seeds. The species' stems can grow between 10 cm and 3 m, hence its ability to form large mantles. The leaves are a bright green (that most of the time stands out from the rest of the present vegetation) and can have a lanceolate or oval shape. They measure between 2.5 and 3.8 centimetres, and both the stem and the leaves have different trichomes distributed over the surface. Ludwigia peploides also have solitary flowers with yellow petals, which measure from one to 1.5 cm in length, and which develop from tassels emerging from the upper part of the axillary bud. The species blooming period occurs from mid-spring to early fall, and the plant is easily identifiable during this period. This is also the period where the species grows the most.

Table 1: DJI P4 Multispectral specifications.

| **Aircraft** | |
|---|---|
| Hover Accuracy Range | RTK enabled and functioning properly: <br> Vertical: ± 0.1 m; Horizontal: ± 0.1 m <br> RTK disabled: <br> Vertical: ± 0.1 m (with vision positioning); <br> ± 0.5 m (with GNSS positioning) <br> Horizontal: ± 0.3 m (with vision positioning); <br> ± 1.5 m (with GNSS positioning) |
| **Camera** | |
| Sensors | Six 1/2.9" CMOS, including one RGB sensor for visible light imaging and five monochrome sensors for multispectral imaging. <br> Each Sensor: Effective pixels 2.08 MP (2.12 MP in total) |
| Filters | Blue (B): 450 nm ± 16 nm; Green (G): 560 nm ± 16 nm; <br> Red (R): 650 nm ± 16 nm; Red edge (RE): 730 nm ± 16 nm; <br> Near-infrared (NIR): 840 nm ± 26 nm |
| Max Image Size | 1600×1300 (4:3.25) |

Table 2: Altitude, time and number of images collected.

| Altitude | Time | Number of images |
|---|---|---|
| 10 m | 11h - 12:45h<br>15:30h - 17h | 435 |
| 15 m | 11h - 12h | 365 |
| 40 m | 10h - 12:45h | 135 |
| 70 m | 11h - 12h | 27 |

### 3.2 Sensor and Platform

Given that our study site is small and the infestation is still at its early stages, we do not need to cover large extents, but we need to have a good resolution. We initially trialled the use of satellite images. However, due to the infestation's low resolution and small coverage area, it was not possible to identify Ludwigia p. on the images. Thus, we decided on using a drone-mounted sensor to build our data set. We used the DJI P4 Multispectral. It is a drone based on the DJI Phantom 4, but the P4 has six sensors instead of a traditional RGB camera. One RGB for visible light and five monochrome sensors for multispectral imaging. Table 1 contains the most relevant drone and camera specifications.

### 3.3 Data Collection

To collect the data for our LudVision data set, we visited the study site twice (October $11^{th}$ and $20^{th}$, 2021). We captured all our data in hover mode because the drone is more stable, allowing us to point the camera straight down.

The data was collected at different altitudes, ranging from 10 m to 70 m. The goal of collecting data from different altitudes is that our model will learn features from a wide range of spatial resolutions. Thus, the final model should be more resilient to altitude variations. The data was also taken at various times to ensure that the model is resilient to variations in solar reflection. Note that no data was collected at solar noon, as it would result in overexposed images due to the light being reflected from the lake's surface. Table 2, contains more detailed information about the collected data.

Every time the drone collects data, it is effectively taking six images, one RGB image in *.jpg* format and five monochromatic images in *.tif* format, one for each band, as can be seen in figure 2. The RGB image should only be used as a reference as it helps to visualize the scene. The remaining five images are monochromatic and used to train the models.

The included 'sunlight sensor' is a feature of the DJI P4 Multispectral that allows the drone to correct the images according to the solar exposure automatically. Usually, this step needs to be done in the pre-processing phase with specialized software, and before each flight, the sensor would need to be calibrated.

### 3.4 Data Pre-processing

Given that the drone already performs corrections to the images based on the light conditions, the only pre-possessing left is to merge the individual bands into a single image. This way, we end up with a five-band image that can be fed to the model for training.

Merging the individual bands is not as straightforward as simply stacking the five bands and exporting them as a final image. The sensors are positioned in a 3×2 array, meaning they all have a unique perspective. Furthermore, we must account for lens imperfections and distortions due to manufacturing tolerances. This way, if we just stacked the bands, the final image would be distorted. Figure 3 contains an example of bands stacked before alignment (the image is distorted and looks overexposed) and bands stacked after alignment (the image is perfectly aligned without distortions).

The alignment was performed using the homography tools from *OpenCV* in *Python*. Once all bands are aligned, they are stacked, and the final image is trimmed to 1400 × 1100 pixels and exported as a TIFF file.

### 3.5 Annotation Process

We use the *Hasty.ai* tool for the annotation process. *Hasty.ai* is an excellent tool for annotating data with organic and complex shapes, which is precisely our case. It has Artificial Intelligence enabled tools that allow for assisted annotation, speeding up the entire process. This tool allowed us to create the ground truth for our dataset, which was used to train and validate the proposed model.

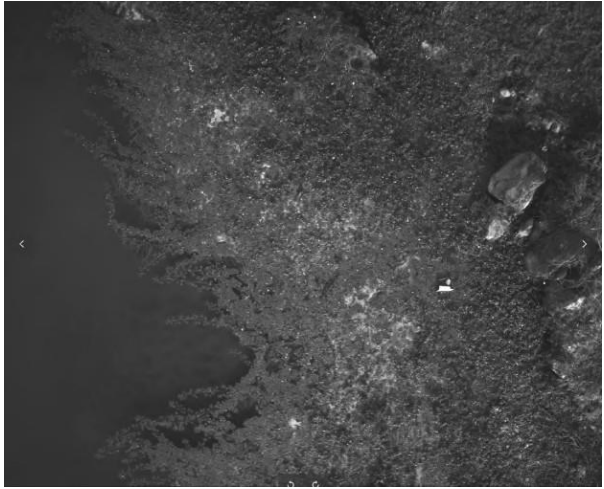
(a)
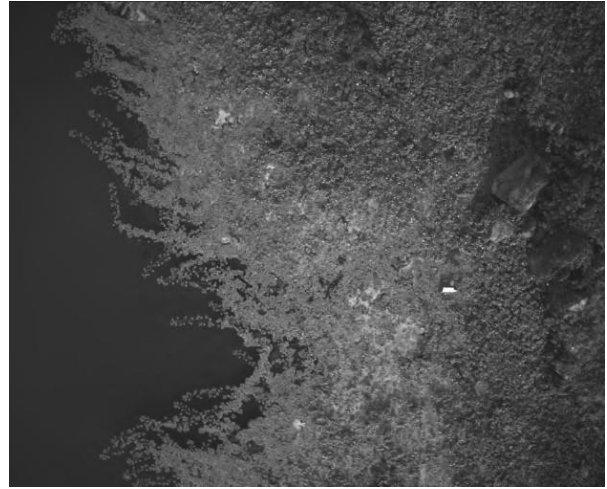
(b)
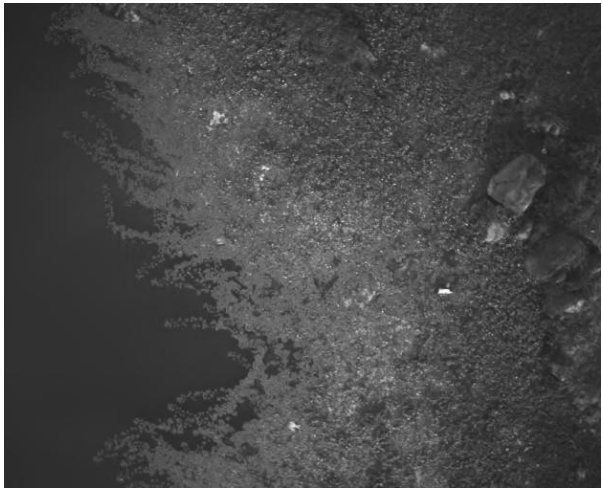
(c)
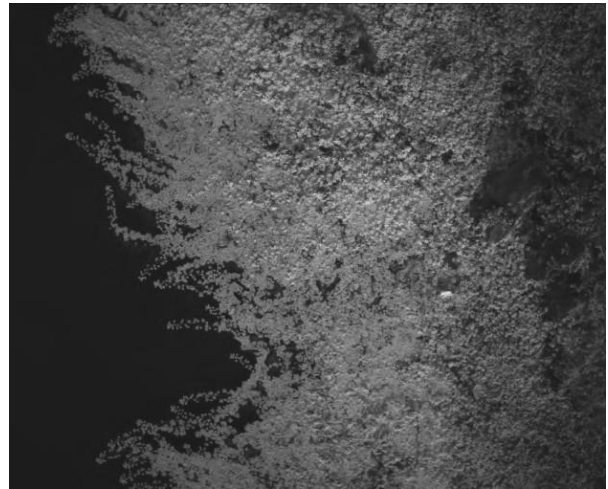
(d)
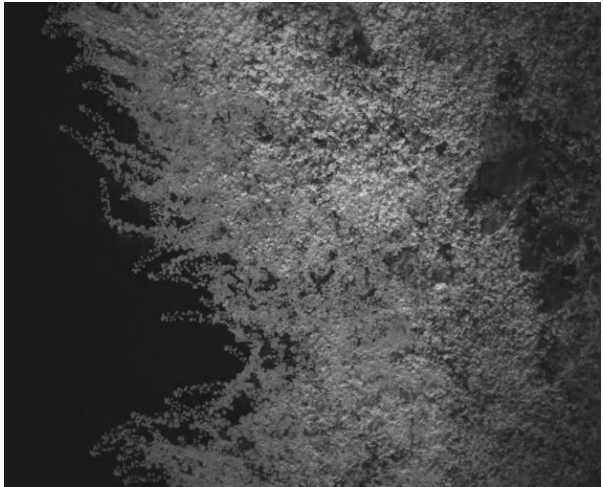
(e)
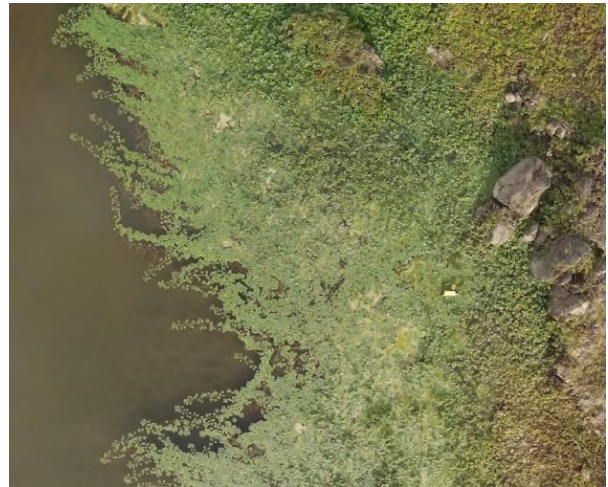
(f)

Figure 2: Example of a red band (a), green band (b), blue band (c), red edge band (d), near infra-red band (e), and RGB image (f) collected by our drone.

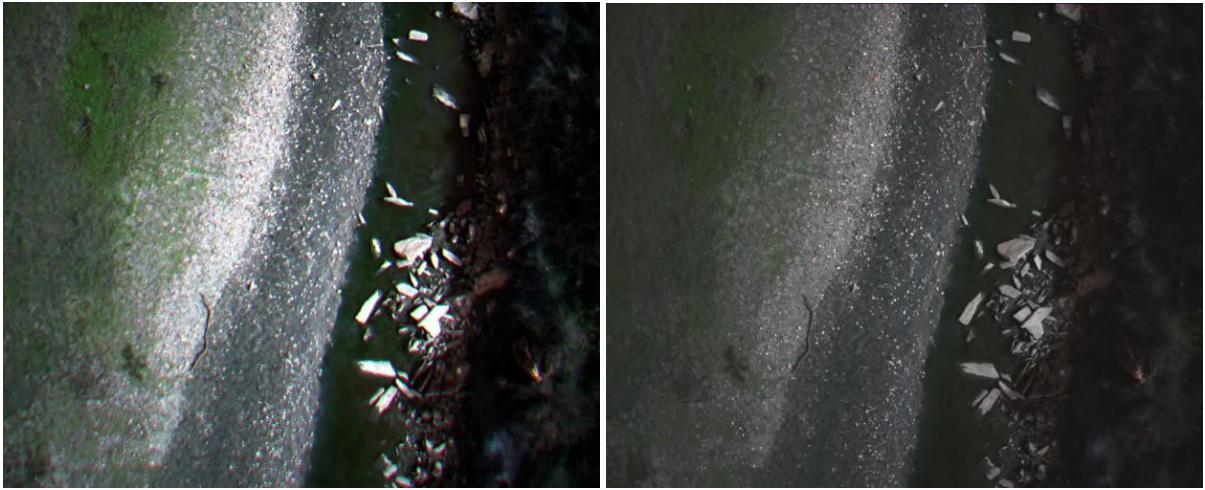

Figure 3: Example of an image created by stacking bands before alignment (a) and stacking bands after alignment (b).

### 3.6 Testing for Spectral Radiance

After completing our data's pre-processing and annotation steps, we assessed the spectral radiance. Our data was captured to allow us to leverage photophysiological measurements. Thus, we measured the radiance for Ludwigia p., rock, surrounding vegetation, and water. As expected, Ludwigia p. has a unique and distinct spectral signature, especially in the non-visible bands (Red Edge and Near Infra-red), as shown in figure 4. This supports our initial argument that a multispectral sensor would be enough to capture significant differences in the spectral radiance, allowing for the detection of Ludwigia peploides.

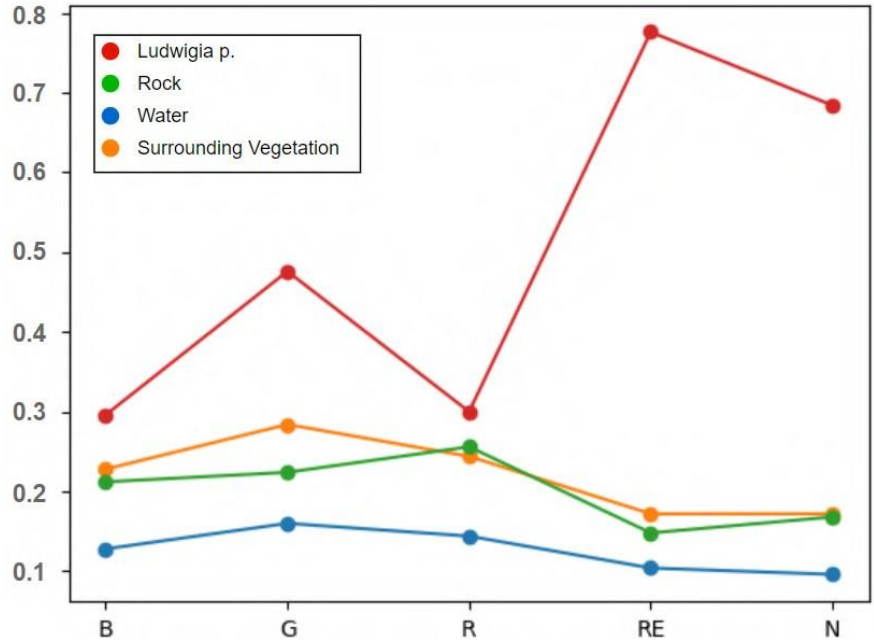

Figure 4: Reflectance values (in %) for each band corresponding to Ludwigia p., water, surrounding vegetation, and rock.

## 4 A New Method for Detection of Ludwigia in Multispectral Images

We analyzed both preliminary concepts regarding RS and state-of-the-art semantic segmentation methods and related work carried out by other authors. After completing our research, we concluded that we should take a "hybrid" approach to our problem by combining both RS concepts and semantic segmentation. The reasoning behind our approach is that despite having a multi-spectral data set, the number of bands in our data set is minute when compared to other studies. We only have five bands, whereas other authors had access to hyperspectral data, with considerably more bands. Thus, we believe that relying solely on the information of only five bands will not be enough to use models like the ones used by other authors [10, 17].

We propose using state-of-the-art semantic segmentation models that will be modified to take our data as input. By using semantic segmentation models, we can leverage both the capabilities of these models to recognize objects and the multi-spectral nature of our data. Fundamentally, the model will have the same behaviour as usual but have access to the information of two additional bands, which should help it find meaningful features to detect the presence of Ludwigia peploides.

We started by performing a series of tests on an existing semantic segmentation model HRNet [24] model in its HRNet + Object Context Recognition (OCR) [23] implementation. Most classification networks, like AlexNet [21] and ResNet [28], gradually reduce the spatial size of the feature maps, connect the convolutions from high-resolution to low-resolution in series, and lead to a low-resolution representation, which is further processed for classification. However, for position-sensitive tasks (e.g., semantic segmentation, object detection, and human pose estimation), high-resolution representations are needed [24]. Wang et al. [24] propose a novel architecture: High-Resolution Net (HRNet), which maintains high-resolution representations throughout the whole process and the learned high-resolution representations are not only semantically strong but also spatially coherent. This is due to two aspects of the network.

First, the authors' approach connects high-to-low resolution convolution streams in parallel rather than in series, which is more common. The way the proposed parallel multi-resolution convolutions work is as follows: The first stage is a high-resolution convolution stream. Then, gradually and one by one, high-to-low resolution streams are added, forming new stages. Lastly, the multi-resolution streams are connected in parallel. As a result, the resolutions for the parallel streams of a later stage consist of the previous stage's resolutions and an extra lower one. This parallel approach allows maintaining the high-resolution, rather than recovering high-resolution from low-resolution, making the representations spatially more precise. Second, the authors repeat multi-resolution fusions to boost the high-resolution representations with the help of the low-resolution representations and vice versa. The goal of the fusion module is to exchange information across multi-resolution representations. As a result, all the high-to-low resolution representations are semantically strong. This fusion scheme is different from most of the other schemes, which aggregate high-resolution low-level and high-level representations obtained by up-sampling low-resolution representations.

Simply put, the network connects high-to-low convolution streams in parallel, maintains high-resolution representations through the whole process, and generates reliable high-resolution representations with strong position sensitivity by repeatedly fusing the representations from multi-resolution streams.

We chose this model for two reasons: First, the model can extract high-resolution representations, which are needed for position-sensitive tasks, like semantic segmentation. Given that the custom fusion module exchanges information across multi-resolution representations, this means that lower-resolution representations also contain information obtained from the high-resolution representations. This allows the model to extract better features for the images at lower resolutions. Given that images taken at high altitudes tend to have low resolution, it makes this model fitting for our data; Second, the mentioned implementation integrates the OCR module [24]. This context module uses a set of pixels lying in the object instead of a set of surrounding sparsely sampled pixels. This is achieved by differentiating the same-object-class contextual pixels from the different-object-class contextual pixels and structuring the contextual pixels into object regions to exploit the relations between pixels and object regions. This allows setting the context for each pixel more accurately. Because Ludwigia p. typically forms large mantles at the water's surface, the module will have a larger area to extract the context for a given pixel. Our data set only has two labels (Ludwigia p. and background), making it easier to differentiate same-object-class contextual pixels from the different-object-class contextual ones.

A few modifications had to be made to train the model on our data, especially on the input layer. Given that the model was originally designed to use traditional RGB imagery, we had to modify the input layer to be able to accommodate our data which has five bands (RGB plus a Near InfraRed and a RedEdge band). The results of these initial tests were used as a baseline to validate our new proposed model.

After establishing a baseline with HRNet [24], we focused on ways we could improve and adapt the model to our data/problem. One of our primary goals is to detect Ludwigia p. at higher altitudes (ideally on satellite imagery). Thus, we tried to find ways to simulate the appearance of satellite data using our data set. When compared to our data, satellite imagery has two distinct differences: it usually has a lower resolution and covers a broader area. So, we had to find ways to both lower our data's resolution and, if possible, enlarge the perceived Field of View (FoV) of the convolution layers.

To achieve the down-sampling of our data, we increased the *stride* value on the fusion models for all connections made between *stage* 1 with the remaining *stages*. This effectively reduces the perceived resolution of our images, making them more closely resemble higher altitude images. To enlarge the perceived FoV, we decided to use *atrous* convolutions (also known as dilated convolutions). This type of convolution allows enlarging the convolution's FoV by increasing the *dilation rate* value. If the *dilation rate* is set to zero, then we effectively have a "conventional" convolution layer. An added benefit of using dilated convolutions is that they save computation and memory costs, as increasing the value of the *dilation rate* allows for a larger receptive field which means viewing more data points.

After implementing the modifications mentioned above, the model is expected to extract low-resolution features, resulting in better performance, especially in data captured at higher altitudes. Its also expected that the model to be more efficient due to the use of dilated convolutions and the down-sampling process.

We performed the following changes to the model:

- We kept the same number of stages, as we did not see substantial benefits in either increasing or decreasing their number;
- We increased the *stride* value on the fusion module from 2 to 3 for all connections made between *stage* 1 with the remaining *stages*. All remaining connections kept the *stride* value at 2.
- We replaced the conventional convolution layers from *stages* 2 and 3, with dilated convolutions, with *dilation* = 2 and *padding* = 2. The convolution layers in *stages* 1 and 4 were kept unchanged.

The architecture of the resulting model is represented in figure 5.

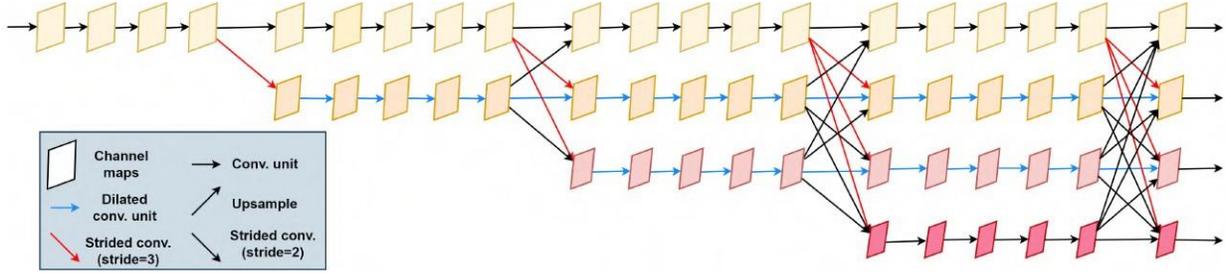

Figure 5: Architecture of our model, that is based on HRNet.

## 5 Results

A series of tests were made to validate our model's performance and compare it to the baseline results. To evaluate our model, we used the following metrics:

- **Producer's accuracy:** is the map accuracy from the point of view of the mapmaker (the producer). This is, how often the real features on the ground are correctly shown on the classified map. It is also the number of reference sites classified accurately, divided by the total number of reference sites for that class.

- **User's accuracy:** is the accuracy from the point of view of a map user. The user's accuracy essentially tells how often the class on the map will be present on the ground. This is referred to as reliability.

- **IoU:** is a metric that determines the extent of overlap between two areas. In the case of semantic segmentation, it is used to determine the overlap between the ground truth and the model's predictions. This metric ranges from 0 to 1 (or 0 to 100%), where 1 means a perfect overlap. We can also calculate the class and mean Intersection over Union (IoU) for multi-class segmentation. As the names suggest, class IoU is calculated by evaluating the overlap for each class, and the mean IoU is the mean value of all classes. Given that we have two classes (*ludwigia* and *background*, we will use class IoU).

Separate training and validation data sets were used. Table 3 contains the results of the best configurations on both HRNet [24] and our proposed model. As can be seen, our model has improved performance compared to HRNet, especially at higher altitudes. Furthermore, we verified that the training times of the model were cut in half. The total training time dropped from 130 h to 57 h (training times of the baseline experiments on HRNet and running the same experiments for validating our model).

If we look closely at the results, we can make the following conclusions. The producer's accuracy has seen an overall increase across all attitudes. Given that producer's accuracy refers to how often the actual features on the ground are correctly shown on the classified map, it means that the classification is more reliable. Thus, when our model classifies a pixel as being Ludwigia p., it is more likely that the prediction is accurate compared to HRNet.

As for the user's accuracy, it decreased for lower altitude images (10/15 m), but it increased for higher altitudes (40 and 70 m). Given that our goal is to classify images at higher altitudes (because it allows us to cover a larger area and possibly upscale our model to satellite images), we consider this a good result. It is essential for us that the model performs better at higher altitudes, even if it means losing some performance at lower altitude classification.

Table 3: Results obtained with HRNet and our proposed model.

|  | HRNet | Our model |
|---|---|---|
| Producer's Acc (10/15 m) | 0.877 | **0.959** |
| Producer's Acc (40 m) | 0.767 | **0.899** |
| Producer's Acc (70 m) | 0.768 | **0.799** |
| User's Acc (10/15 m) | **0.925** | 0.850 |
| User's Acc (40 m) | 0.861 | **0.916** |
| User's Acc (70 m) | 0.850 | **0.955** |
| Class IoU (10/15 m) | **0.835** | 0.820 |
| Class IoU (40 m) | 0.749 | **0.830** |
| Class IoU (70 m) | 0.720 | **0.769** |

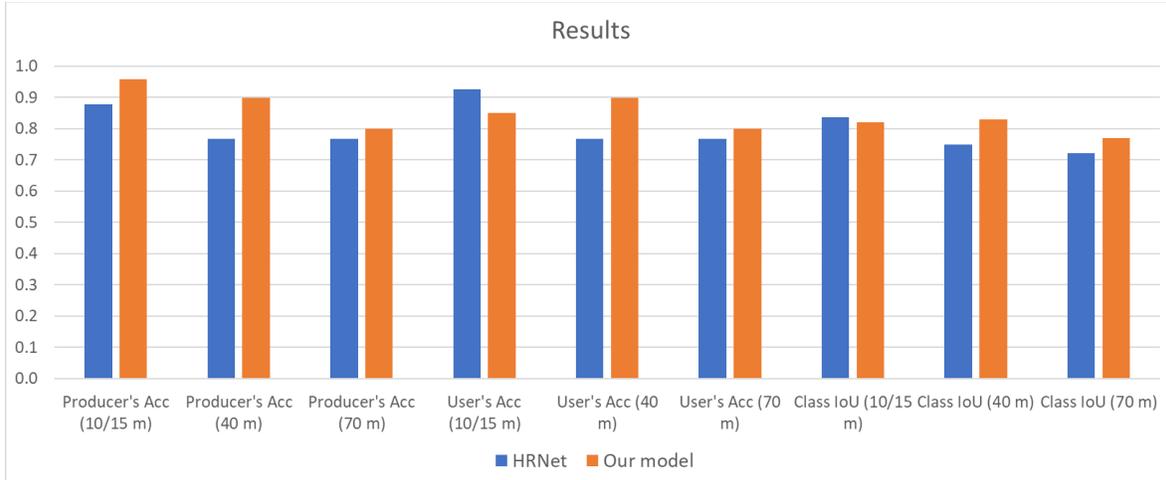

Figure 6: Performance comparison between HRNet and our model

Class IoU, follows the same trend as the user's accuracy. Our model has slightly lower performance at low altitude images, but it outperforms HRNet at higher altitudes.

## 6 Discussion

### 6.1 Model Validation

Due to the modifications, we made to the original HRNet, our new model is better fitted for our application. This means that there is no need to create a custom model from scratch. It is possible to create a new semantic segmentation method by modifying existing models so that the new model is better suited for a given application. This allows reducing the time needed to develop said, custom model.

Analyzing all the metrics in general, we conclude that it outperforms the HRNet model, which already demonstrated excellent results in correctly classifying Ludwigia peploides. Our model's gain in performance is especially noticeable in high-altitude images, which is a desirable outcome. Our proposed model is able to detect the presence of the targeted species with great accuracy.

Furthermore, we verified that the training times of the model were cut in half. This not only means that our model is more capable of identifying Ludwigia p., but it is also faster to deploy due to the shorter training time.

By looking at both the metrics and model output, we can conclude that our model produces high-reliability maps and has the tendency to output more false positives than false negatives. It is better to have a model that occasionally identifies other species as Ludwigia p. than to have a model that sometimes fails to identify the IAS. This is crucial because in real-world scenarios, not identifying a single strand of Ludwigia p. may lead to identifying a site falsely as not being infested. Due to the capability of the IAS spreading rapidly from one single seed, this can lead to a severe infestation in the future. Ludwigia p. reproduces aggressively, and one misclassified plant can lead to a wholly covered body of water in a brief period.

### 6.2 Analyzing the Model's Output

After analyzing and comparing quantitative results, we shifted our focus to analyzing the actual outputs of the model. Although evaluating the model using quantitative results is imperative, a visual analysis of the predictions allows us to understand the scenarios where the model is and is not accurate.

Considering the quantitative results and our observations, we further conclude that the model performs well. As seen in figures 7 the model's predictions are very close to the respective ground truths. Figure 8 presents the overlap of the model's predictions on the respective images, where it is clear that the model is able to correctly and reliably identify the targeted species at multiple altitudes.

### 6.3 Comparing Our Results to Similar Studies

To further validate our model, we compared its performance with similar studies. We compare our model's performance to the performance achieved by Bolch in [10]. His work, setup, and objectives are very similar to ours, and he also assesses the performance of water primrose (Ludwigia spp), which is very similar to Ludwigia p.). He achieved a producer's accuracy of 100% on water primrose and a user's accuracy of just 50%. Our model at 70 m (the highest altitude on our data set), achieved a producer's accuracy of 79.9% (a 20.1% decrease compared to [10]), and a user's accuracy of 95.5% (an increase of 45.5%). We have to keep in mind that the sensor used in [10] is superior to ours and has more bands (their sensor has 270 bands, while our sensor only has five bands). This further proves not only our method's validity but also that great classification results can be achieved using data sets built with data captured from general application sensors.

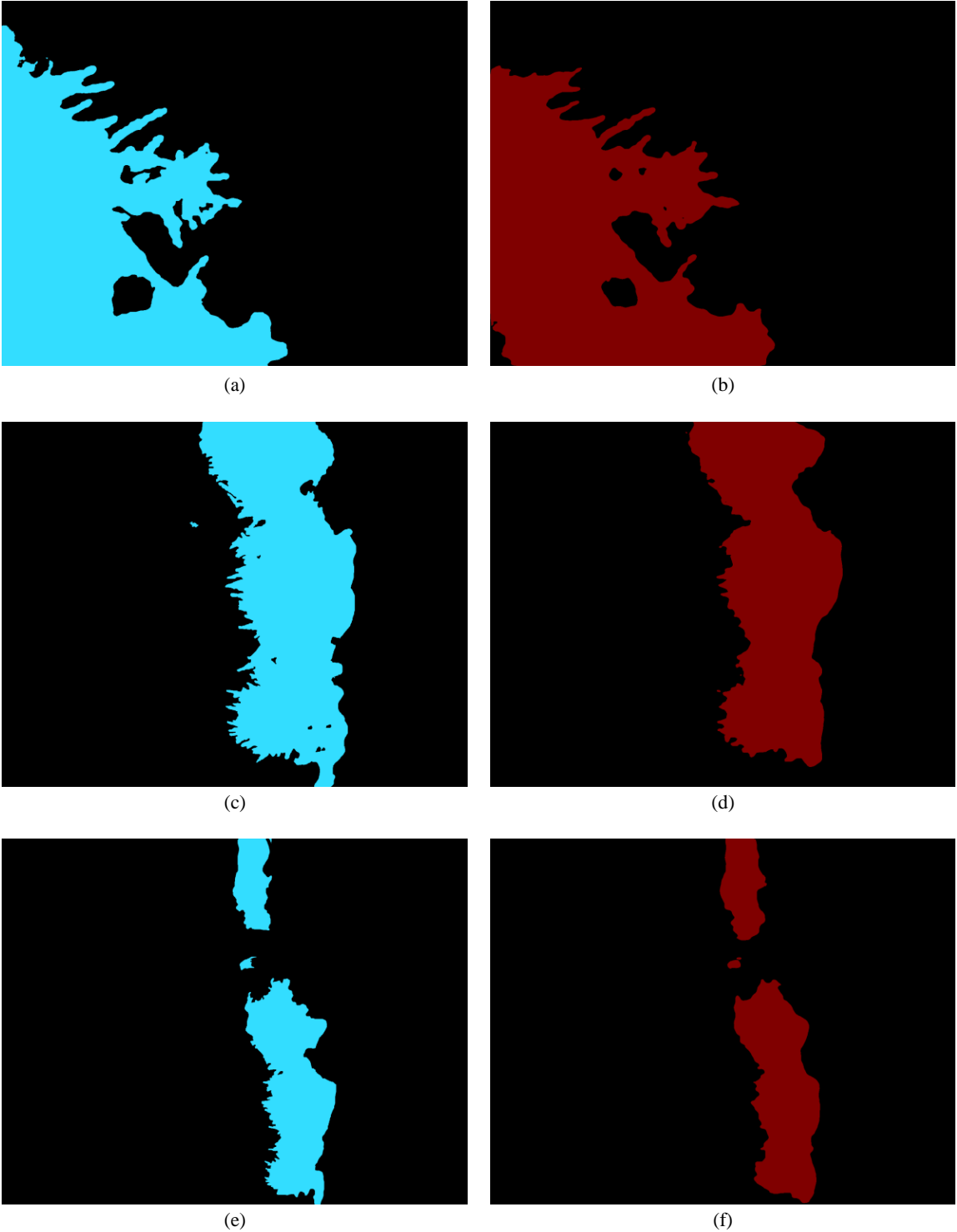

Figure 7: Ground truths and their respective model predictions. Images (a) and (b) are an example of the ground-truth and model predictions corresponding to images taken at 10/15 m, (c) and (d) at 40 m, and (e) and (f) at 70 m.

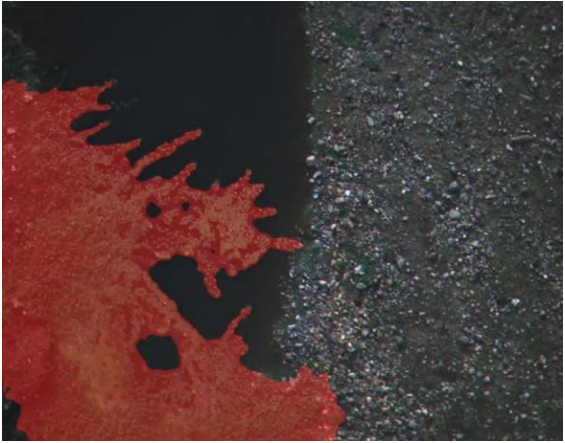

(a)

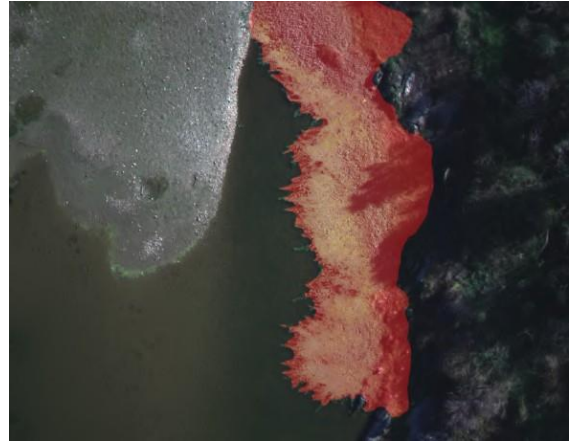

(b)

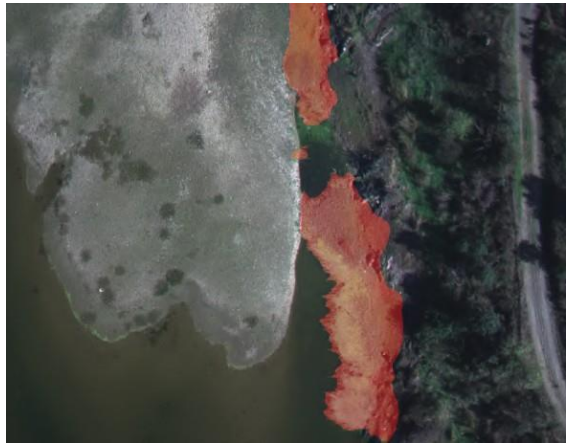

(c)

Figure 8: Predictions overlapped on their respective images.

# 7 Conclusion

The detection of IAS is a very pertinent topic. We set out to accomplish this type of detection using a low-cost sensor and created a new data set that is freely available. We also created a new approach to the detection of Ludwigia. p. by modification of a semantic segmentation method. We showed how segmentation models could be used to identify the targeted species. By using semantic segmentation models, we were able to leverage both the capabilities of these models to recognize objects and the multispectral nature of our data. Before being able to start training and testing models, we need first to have a data set. Unfortunately, we could not find any publicly available data sets and had to build our own. We captured data at multiple altitudes, periods of the day, and atmospheric conditions. We wanted to make sure that our data set was as representative as possible and contained variations of the atmospheric conditions. After collecting the necessary images, we started the processing and annotation steps of the data set. Once annotated, we could use our data set to train and test semantic segmentation models. We first used a preexisting model (HRNet [24]), which was modified to handle our data. Once we validated the concept of using a semantic segmentation model to detect Ludwigia p., we established a baseline. We proposed a set of modifications to the HRNet model to increase its performance, especially at high altitudes, and reduce the training time. The proposed model needs considerably less training time while increasing performance in high-altitude scenarios. Comparing our results with the results other authors achieved, the proposed method has comparable performance using only five image channels as opposed to several dozens or hundreds of data channels.

## Supplementary Materials

The complementary academic thesis of this work will be available at `http://ludvision.di.ubi.pt`.

## Author Contributions

**António Abreu:** Conceptualization, Data collection planning and implementation, Data processing and annotation, Model development, Experimentation and validation, Data analysis and interpretation, Validation, Visualization, Led manuscript writing with primary contributions from Luís A. Alexandre. **Luís A. Alexandre:** Funding acquisition, Conceptualization, Supervision, Writing – reviewing & editing, Visualization. **João Santos:** Conceptualization, Supervision, Data collection planning and implementation. **Filippo Basso** Conceptualization, Supervision, Data collection planning and implementation, Funding.

## Funding

This research was funded by Zirak srl – Information Technology and by NOVA LINCS (UIDB/04516/2020) with the financial support of FCT-Fundação para a Ciência e a Tecnologia, through national funds.

## Data Availability Statement

The data presented in this study will be available at `http://ludvision.di.ubi.pt`.

## Acknowledgments

The authors would like to thank Hasty.ai for providing a subscription, free of charge, that allowed to fully access all the features. It made the annotation process easier and faster, allowing the authors more time to focus on more critical aspects of this work.

## Conflicts of Interest

The authors declare no conflict of interest.